\documentclass[journal,twoside,web]{IEEEtran}
\usepackage{tmi}
\usepackage{cite}
\usepackage{amsmath,amssymb,amsfonts}
\usepackage{graphicx}
\usepackage{textcomp}
\usepackage{latexsym}
\usepackage{cancel}
\usepackage{booktabs}
\usepackage{algorithm}
\usepackage{algpseudocode}
\usepackage{algorithmicx}
\usepackage{soul}
\usepackage{hyphenat}
\usepackage{microtype}
\usepackage{hyperref}
\algdef{SE}[DOWHILE]{Do}{doWhile}{\algorithmicdo}[1]{\algorithmicwhile\ #1}%

\usepackage{url}
\def\BibTeX{{\rm B\kern-.05em{\sc i\kern-.025em b}\kern-.08em
    T\kern-.1667em\lower.7ex\hbox{E}\kern-.125emX}}
\begin{document}
\title{BayTTA: Uncertainty-aware medical image classification with optimized test-time augmentation using Bayesian model averaging}

\author{Zeinab Sherkatghanad,\thanks{Z. Sherkatghanad and M. Bakhtyari are with the Department of Computer Science, Université du Québec à Montréal, H2X 3Y7, Montreal, QC, Canada} Moloud Abdar, \IEEEmembership{Member, IEEE}, \thanks{M. Abdar is with the Institute for Intelligent Systems Research and Innovation (IISRI), Deakin University, 3216, Geelong, VIC, Australia (m.abdar1987@gmail.com)} Mohammadreza Bakhtyari, Paweł Pławiak, \thanks{P. Pławiak is with the Department of Computer Science, Faculty of Computer Science and Telecommunications, Cracow University of Technology, Krakow, Poland} Vladimir Makarenkov \thanks{V. Makarenkov is with the Department of Computer Science, Université du Québec à Montréal, H2X 3Y7, Montreal, QC, Canada and also the Mila - Quebec AI Institute, Montreal, QC, Canada.}}

\maketitle

\begin{abstract}
Test-time augmentation (TTA) is a well-known technique employed during the testing phase of computer vision tasks. It involves aggregating multiple augmented versions of input data. Combining predictions using a simple average formulation is a common and straightforward approach after performing TTA. This paper introduces a novel framework for optimizing TTA, called BayTTA (Bayesian-based TTA), which is based on Bayesian Model Averaging (BMA). First, we generate a prediction list associated with different variations of the input data created through TTA. Then, we use BMA to combine predictions weighted by the respective posterior probabilities. Such an approach allows one to take into account model uncertainty, and thus to enhance the predictive performance of the related machine learning or deep learning model. We evaluate the performance of BayTTA on various public data, including three medical image datasets comprising skin cancer, breast cancer, and chest X-ray images and two well-known gene editing datasets, CRISPOR and GUIDE-seq. Our experimental results indicate that BayTTA can be effectively integrated into state-of-the-art deep learning models used in medical image analysis as well as into some popular pre-trained CNN models such as VGG-16, MobileNetV2, DenseNet201, ResNet152V2, and InceptionRes-NetV2, leading to the enhancement in their accuracy and robustness performance.
%%%% enhancing the robustness and accuracy of machine learning models compared to simple averages across different datasets.
The source code of the proposed BayTTA method is freely available at: \underline
{https://github.com/Z-Sherkat/BayTTA}.
\end{abstract}


\begin{figure*}[t]
\centering
\includegraphics[scale=.5]{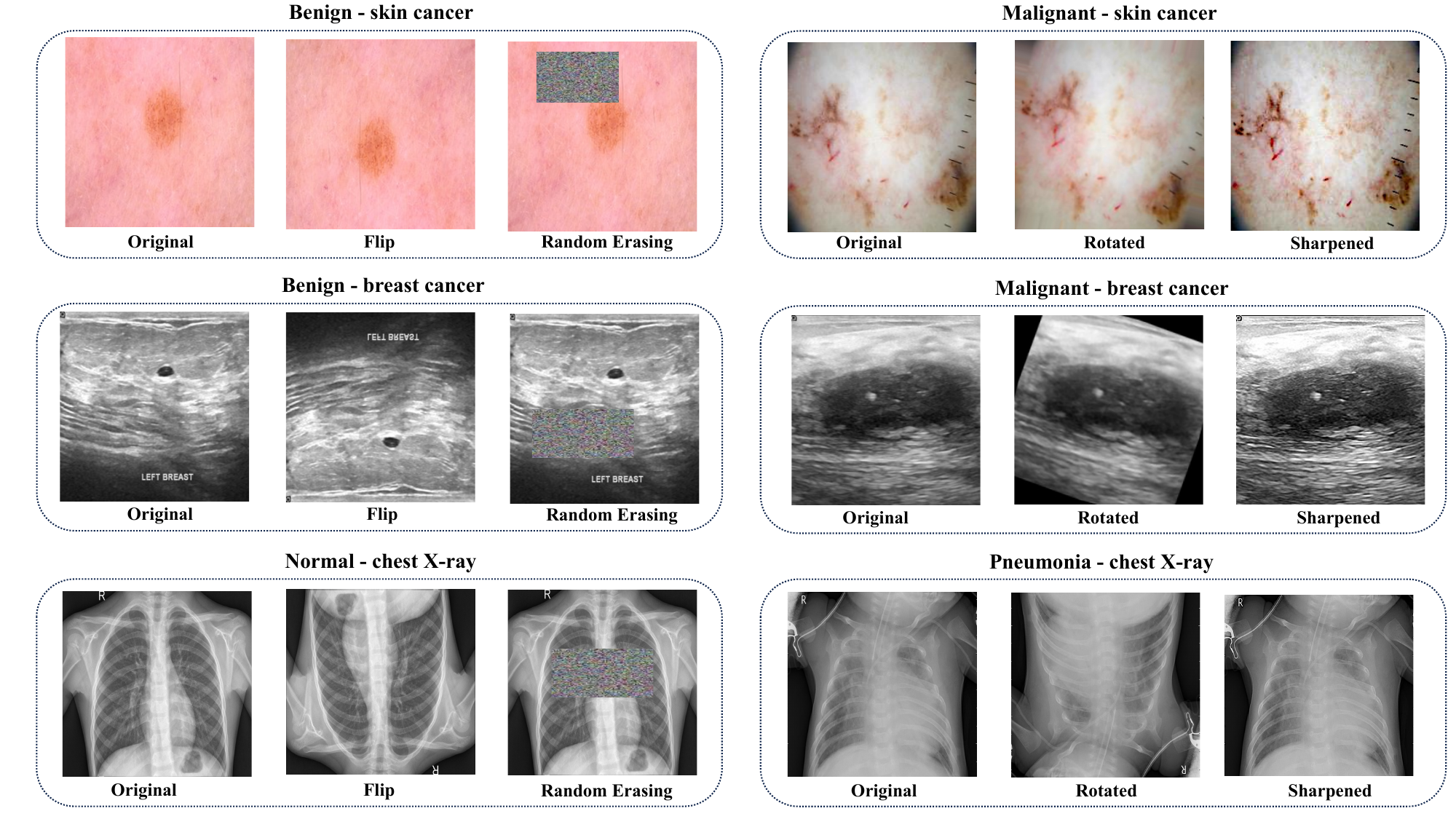}
\caption{Examples of original benign and malignant skin cancer, breast cancer, and chest X-ray images and their augmented versions considered in our study.}\label{fig1}
\end{figure*}
%% main text
\section{Introduction}
\label{sec:introduction}
\indent Deep learning models are highly effective tools for identifying complex patterns in data. However, they are also prone to overfitting and memorizing the training data rather than generalizing real data patterns. Data augmentation is a critical strategy to overcome this issue, which involves synthesizing new training samples by applying to the original data a series of transformations and perturbations, such as random erasing, flipping, rotating, and scaling (see Fig. \ref{fig1}). Leveraging data augmentation techniques effectively introduces variability in the training data, leading to a more robust and general model. Machine learning (ML) and deep learning (DL) models equipped with data augmentation can handle the complexities inherent in medical image data, often leading to an enhanced diagnostic accuracy \cite{tsuneki2022deep, biswas2023data, goceri2023medical, bozkurt2023skin}. \\ 
\indent Test-Time Augmentation (TTA) is a popular data augmentation technique which can increase the prediction power of ML and DL models. TTA involves applying data augmentation on test input rather than on training data. It provides a more reliable estimate of the target variable by averaging the predictions across augmented inputs to find the output result. While data augmentation is mainly used for improving training data diversity to enhance the model generalization on unseen data, TTA produces multiple versions of each test image to obtain a more robust estimate of the target variable \cite{jin2018deep, maron2021robustness}.\\
\indent Recently, Uncertainty Quantification (UQ) methods have been effectively applied in a variety of research domains, contributing to the robustness and reliability of the optimization and decision-making processes. There are two primary categories of uncertainty \cite{abdar2021review, hullermeier2021aleatoric}, referred to as aleatoric and epistemic. While aleatoric uncertainty pertains to the inherent characteristic of the data distribution, epistemic uncertainty relates to the lack of knowledge about the optimal model. Uncertainty quantification is a vital tool for addressing the limitations of data (aleatoric uncertainty) and models (epistemic uncertainties) in various scientific and engineering applications, and thus for improving the trustworthiness of predictions, especially in the testing phase \cite{abdar2021uncertainty}. Although deep learning models often outperform traditional machine learning methods, overconfident predictions remain their crucial issue, leading to excessive prediction errors \cite{sherkatghanad2020automated}. By combining uncertainty quantification with TTA, we can gain a deeper insight into the reliability and robustness of model predictions \cite{wang2019aleatoric, bahat2020classification, ayhan2022test}. This is particularly beneficial for machine learning and deep learning applications in healthcare, autonomous systems, and safety-critical domains, where understanding and managing uncertainty may be critical.\\
\indent Here, we describe a novel technique, called BayTTA, that employs Bayesian Model Averaging (BMA) logistic regression to optimize TTA. Our new method applies BMA on the predictions obtained through TTA. BMA can effectively handle model uncertainty, combining multiple models generated by TTA and offering precise and robust predictions. It aggregates predictions from all candidate models, weighted by their posterior probabilities, and results in a model-averaged prediction that accounts for model uncertainty. \\
This article showcases a number of contributions that are as follows:
\begin{itemize}
\item We propose the BayTTA method for optimizing TTA using the BMA approach and apply it to medical image classification. 
\item We estimate the uncertainty associated with predictions obtained through TTA and BMA, offering further insight into the reliability of our method.
\item We assess the performance of BayTTA on three public medical image datasets, including skin cancer, breast cancer, and chest X-ray images available on the Kaggle and Mendeley data repositories as well as on two popular gene editing datasets, CRISPOR and GUIDE-seq, generated using the CRISPR-Cas9 technology \cite{haeussler2016evaluation}. We demonstrate that our method can be applied successfully with different deep learning network architectures. Our results indicates a superior performance of the proposed method compared to standard averaging.
\item We evaluate the performance of BayTTA used in combination with some well-known pre-trained deep learning networks, including VGG-16, MobileNetV2, DenseNet201, ResNet152V2, and InceptionResNetV2. 
\item We demonstrate how incorporating BayTTA enhances the predictive power and robustness of state-of-the-art deep learning models used in medical image analysis, compared to standard TTA. 
%The integration of TTA data augmentation through BMA showcases the effectiveness of our approach.
\end{itemize}

\section{Related work}
\subsection{Uncertainty quantification}
{\bf{Technical progress:}} Uncertainty quantification (UQ) is essential to enhance the credibility of predictions during the testing phase. Since the excessive confidence of deep neural networks may lead to prediction errors, it is imperative to address the issue of overconfident predictions in order to improve their reliability and trustworthiness \cite{hamedani2023breast}. As deep learning models are now constantly used in critical areas, the ability to quantify and manage uncertainty becomes increasingly vital \cite{hoffmann2021uncertainty, mazoure2022dunescan, abdar2023uncertaintyfusenet}.

Nowadays, UQ has important applications in image processing, computer vision, medical image analysis, diagnostic modeling, and healthcare decision-making \cite{abdar2021review, lambert2022trustworthy, loftus2022uncertainty, seoni2023application}. There are several well-known methods for measuring uncertainty such as Monte Carlo dropout \cite{gal2016dropout}, Variational Inference \cite{wang2020doubly,rudner2022tractable}, Deep Ensembles \cite{d2021repulsive, rahaman2021uncertainty, abe2022deep}, and Bayesian Deep Ensembles \cite{he2020bayesian}. Bayesian Model Averaging (BMA) is another effective technique used to take into account prediction uncertainty \cite{wintle2003use, monteith2011turning, izmailov2021dangers, bartovs2021bayesian}. Development of BMA in the context of model uncertainty has been influenced by the seminal works of \cite{draper1995assessment, draper2013bayesian} and \cite{hoeting1998bayesian} as their original methods provide insight into the quantification of both epistemic and aleatoric uncertainties.

{\bf{UQ application in medical image analysis:}} Nowadays, UQ methods have become a tool of choice for estimating the uncertainty associated with disease detection, diagnosis, medical image segmentation, and identification of the region of interest (ROI) in medical image analysis. In their recent work, \cite{abdar2021uncertainty} have introduced an efficient hybrid dynamic model of uncertainty quantification, called TWDBDL. The model is based on the Three-Way Decision (TWD) theory and Bayesian Deep Learning (BDL) methods used together to improve the trustworthiness of predictions in skin cancer detection. Edupuganti et al. \cite{Edupuganti2021} used a probabilistic variational autoencoders (VAEs), i.e. a Monte Carlo technique to generate pixel uncertainty maps, and Stein’s Unbiased Risk Estimator (URE) to provide accurate uncertainty estimations in knee magnetic resonance imaging. Gour et al. \cite{gour2022uncertainty} introduced the UA-ConvNet, i.e. an uncertainty-aware Convolutional Neural Network (CNN) model, for COVID-19 detection in chest X-ray (CXR) images. The model estimates the uncertainty based on the EfficientNet-B3 Bayesian network supplemented with Monte Carlo dropout.

Mazoure et al. \cite{mazoure2022dunescan} designed a novel web server, called DUNEScan, for uncertainty estimation in CNN models applied to skin cancer detection. The web server employs binary dropout to compare the average model predictions, providing visualization for uncertainty to diagnose precisely skin cancer cases. Han et al. \cite{Han2024} proposed a novel dynamic multi-scale convolutional neural network (DM-CNN) that leverages a hierarchical dynamic uncertainty quantification attention (HDUQ-Attention) submodel. HDUQ-Attention includes a tuning block for adjusting the attention weights as well as Monte Carlo dropout for quantifying uncertainty. The experiments conducted on skin disease images (HAM10000), colorectal cancer images (NCT-CRC-HE-100K), and lung disease images (OCT2017 and Chest X-ray) demonstrated that the DM-CNN model accurately quantifies uncertainty, showing a stable performance.
 
{\bf{UQ application in gene editing:}} Uncertainty quantification can be effectively applied in the context of gene editing in order to improve the trustworthiness of on- and off-target predictions in CRISPR-Cas9 experiments \cite{sherkatghanad2023using}. 

For example, Zhang et al. \cite{zhang2020dl} presented a deep learning model for off-target activity prediction, employing data augmentation to mitigate the class imbalance issue. The authors collected data from two source types, i.e. in vitro and cell-based experiments, to increase the size of the positive class samples (off-targets). They suggested synthetically expanding the number of positive samples by rotating the sgRNA-DNA encoded images by 90, 180, and 270 degrees, respectively, to enhance the model competency.

Moreover, Kirillov et al. \cite{kirillov2022uncertainty} have recently introduced a pioneering method that incorporates uncertainty into off-target predictions. Their approach offers an interpretable evaluation of Cas9-gRNA and Cas12a-gRNA specificity through deep kernel learning, estimating a gRNA's cleavage efficiency with a corresponding confidence interval.

\subsection{Test-time augmentation}
\begin{figure}[!t]
\centering
\includegraphics[scale=.45, angle=270]{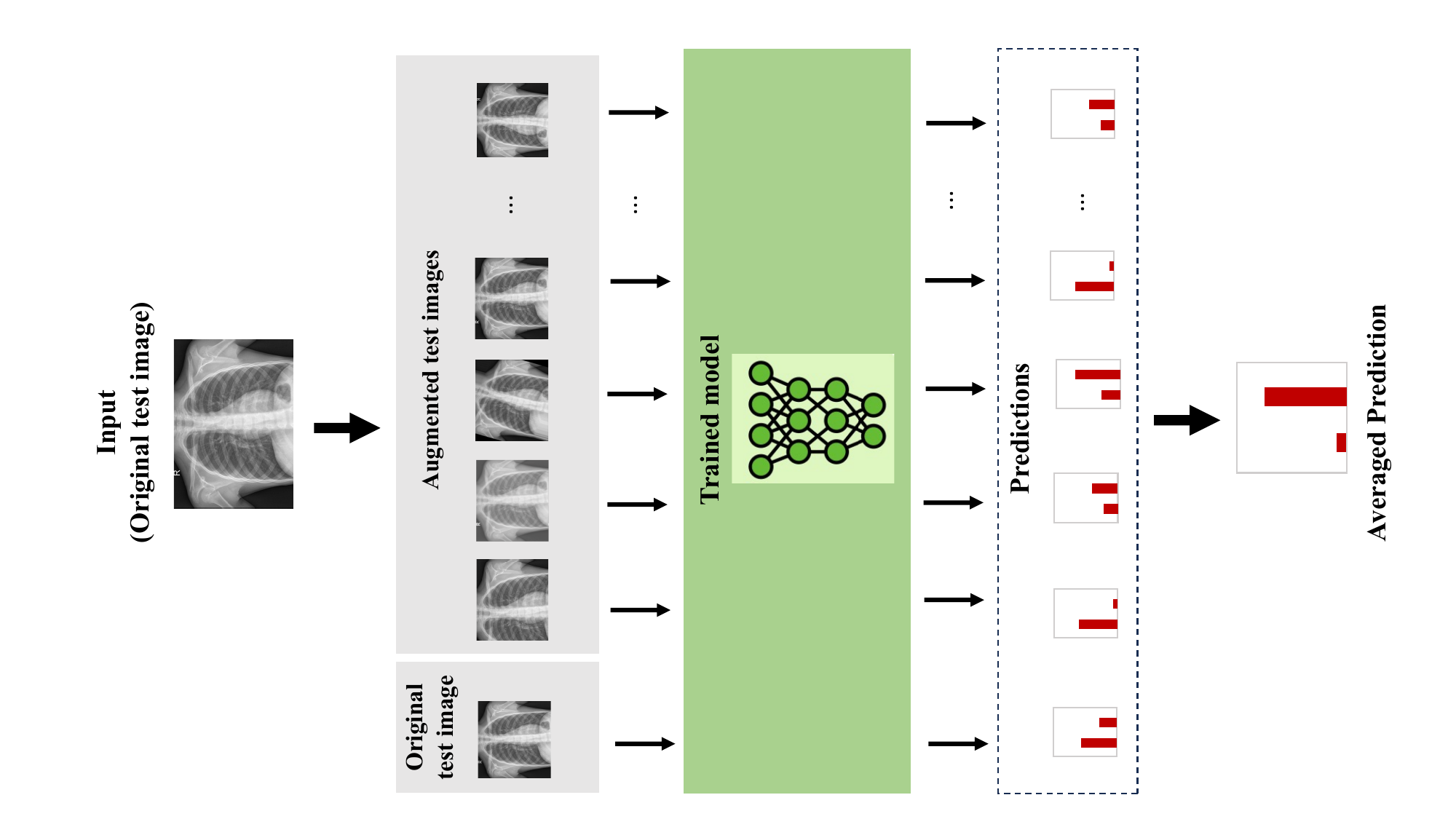}
\caption{A schematic view of a conventional test-time augmentation (TTA) process.}\label{fig2}
\end{figure}

{\bf{Technical progress:}} Test-time augmentation (TTA) is a well-known technique that applies data augmentation during the testing phase. TTA has multiple benefits, including improving the model generalization and reliability capacity \cite{song2017pixeldefend, cohen2019certified}, estimating uncertainty in model predictions \cite{wang2019aleatoric, bahat2020classification, ayhan2022test}, and boosting accuracy in classification and segmentation tasks \cite{krizhevsky2012imagenet, szegedy2015going, he2016deep, shanmugam2021better}. Fig. \ref{fig2} presents a detailed flowchart of a typical TTA process.

Recently, some advanced TTA methods using diverse aggregation strategies have been proposed. They include, among others, selective augmentation techniques such as the instance-aware TTA based on a loss predictor \cite{kim2020learning}, the instance-level TTA with entropy weight method \cite{chun2022cyclic}, and the selective-TTA method \cite{son2023efficient}.

\cite{lyzhov2020greedy} presented a straightforward and efficient method based on a policy search algorithm, known as greedy policy search (GPS), designed to optimize image classification. The method aims to optimize and determine the most effective data augmentation strategy to be applied during the testing phase of a machine learning process, and thus to improve the prediction accuracy and robustness of ML models. Kim et al. \cite{kim2020learning} developed an instance-aware test-time augmentation approach that employs a loss predictor to dynamically select test-time transformations based on the expected losses for individual instances. Furthermore, Chun et al. \cite{chun2022cyclic} introduced an instance-level TTA with Entropy Weight Method (EWM) as an innovative approach to improve the accuracy and robustness of classification models.

The paper of D. Shanmugam \cite{shanmugam2021better} describes a new method for aggregating model predictions obtained from TTA. Unlike the traditional approach of averaging model predictions, this method focuses on learning different augmentation weights to aggregate predictions obtained from transformations during TTA. The authors noticed that existing aggregation methods, based on the mean or the maximum of predictions obtained from augmented images, may not be optimal because they do not consider the relationship between the original test image and its augmented versions. The paper also offers insights into the point when TTA can be helpful and provides guidance regarding the use of different TTA policies. Moreover, Shanmugam and co-authors characterized the cases where TTA transforms correct predictions into incorrect ones, and vice versa.

{\bf{TTA application in medical image analysis:}} TTA has been extensively studied by the medical imaging community due to its capability to contribute to model robustness and improve the trustworthiness and generalization of predictions during the testing phase. Wang et al. \cite{wang2019aleatoric} conducted critical research in the context of deep convolutional neural network (CNN)-based medical image segmentation. Their work focuses on epistemic and aleatoric uncertainty analysis at pixel and structure levels, providing valuable insights into the reliability of segmentation results. Gaillochet et al. \cite{gaillochet2022taal} introduced a simple and powerful task-agnostic semi-supervised active learning segmentation approach, called TAAL (TTA for Active Learning). TAAL exploits unlabeled samples during training and sampling phases by using a technique known as cross-augmentation consistency.

\section{Methodology}
\begin{figure*}[!t]
\centering
\includegraphics[scale=.53]{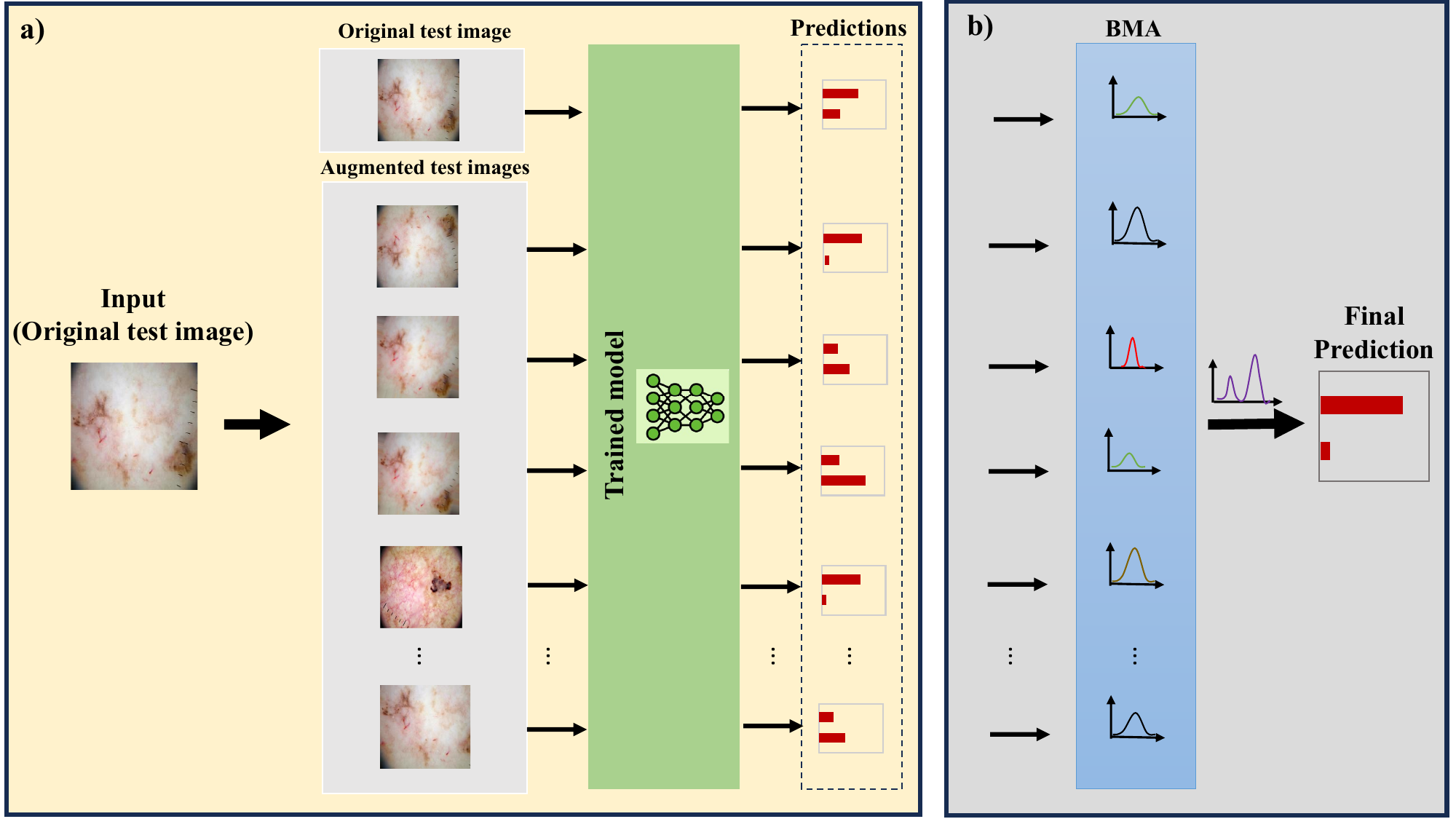}
\caption{An overview of the proposed BayTTA method. During the testing phase: (a) TTA generates predictions from a set of fixed augmented images, and (b) BMA is then applied to combine and aggregate these predictions by treating each unique combination as a distinct candidate model.}\label{fig3}
\end{figure*}

In this section, we elaborate on the background, foundation, and comprehensive explanation of the methodological procedures used within the proposed method. Our new method, called BayTTA, aims to optimize TTA using BMA and uncertainty estimation. 
%The model structure comprises these two integral components within a consistent mathematical framework.
The first component of our method involves formulating a mathematical model that applies BMA to output predictions generated from multiple transformed versions of the input data (e.g. input images). The second component of the method estimates the uncertainty of the predictions obtained from an augmented set of test images associated with image transformations or noise. Together, these components offer a comprehensive approach to enhance the robustness and reliability of ML or DL models during the testing phase. An overview of the proposed BayTTA method is presented in Fig. \ref{fig3}.

%\subsection{Test-time augmentation}
%\subsection{Aggregation via BMA}
Traditional TTA involves applying a number of data augmentation techniques, including rotation, cropping, flipping, and brightness adjustments, to the test data before making predictions. This widely used approach leverages ensemble methods, such as averaging or taking the majority vote of predictions made on augmented samples, to reduce the impact of random noise and variation in the test data. TTA usually produces stable and accurate predictions by averaging multiple augmented versions of the test dataset \cite{wang2019aleatoric, wang2019automatic}.
%TTA is a simple, efficient, and powerful technique for enhancing model performance and uncertainty quantification, especially in challenging and dynamic real-world scenarios.
%This research showcases an advanced TTA technique that has evolved to tackle specific challenges and enhance model performance across various domains. Our findings reveal that incorporating BMA within a standard TTA framework can optimize TTA and offer a sound approach to combine predictions from multiple model instances, thereby boosting the effectiveness of TTA.

%\begin{figure*}[!t]
%\vspace{-30mm}
%\centering
%\includegraphics[scale=.7, angle=270]{FLOWCHART.pdf}
%\caption{Flowchart of the proposed BayTTA method.}\label{fig4}
%\end{figure*}

Our study implements BMA to integrate predictions obtained from TTA (i.e. predictions obtained from the same model but using different distorted images), thus deviating from the conventional averaging approach used in TTA to yield more precise and robust outcomes. Let us focus on a single image $I_y$ from the test dataset, where \( y \in \{0, 1\}\) denotes its classification label. In a scenario generating \(k\) augmented versions of the test image in TTA, the input to the  BMA comes from the vector \({\mathbf{x} = \{x_1, x_2, ..., x_{k+1} \}}\), where each element corresponds to a prediction from the TTA procedure. Here, \(x_1\) denotes the prediction derived from the original test image $I_y$, and \(\{x_2, x_3, ..., x_{k+1}\}\) denotes the predictions from augmented versions of this image.

% Our study also implements BMA to integrate {\color{red} 'prediction list' where the predictions obtained from the same model but with different distorted versions of the image,} thus deviating from the conventional averaging approach used in TTA, to yield more precise and robust outcomes. Let us focus on a single image $I_y$ from the test dataset, where \( y \in \{0, 1\}\) denotes its classification label. In a scenario generating \(k\) augmented versions of the test image in TTA, the input to the  BMA comes from the vector \({\mathbf{x} = \{x_1, x_2, ..., x_{k+1} \}}\), where each element corresponds to a prediction from the TTA procedure. Here, \(x_1\) denotes the predictor variable derived from the original test image $I_y$ and \(\{x_2, x_3, ..., x_{k+1}\}\) denote the predictor variables from augmented versions of this image.

Rather than treating the TTA outputs as independent predictions to be averaged, we consider them as predictor variables, each with a unique combination, defining a distinct candidate model within the BMA framework. BMA involves generating various models by combining predictor variables and selecting candidate models based on likelihood. To formalize this process, we introduce \(\mathcal{P(\mathbf{x})}\), a non-empty power set of \(\mathbf{x}\), representing all combinations of predictor variables. Subsequently, utilizing \(I \ \in \mathcal{P(\mathbf{x})}\), we select desired predictor variables from \(\mathbf{x}\) for model \(M_I\), where \(\mathbf{x}_I\) represents the input of the respective model \(M_I\). %Following this, we define a threshold \(\mathcal{L}_{max}\), representing the maximum likelihood value among the observed models, for selecting candidate models based on likelihood. 
In the context of Bayesian model averaging logistic regression, we further elucidate the mathematical formulations that define our algorithm in the subsequent discussion.

According to the Bayes theorem, the posterior distribution for model \(M_I\) is given by:
\begin{equation}\label{eq1}
p(M_I\mid\mathbf{x}_I,y) = \frac{p(\mathbf{x}_I,y \mid M_I)p(M_I)}{\mathcal{L}_{total}},
\end{equation}
where \(p(M_I)\) denotes the prior probability of \(M_I\), \(\mathcal{L}_{total}\) is the marginal likelihood, and  $p(\mathbf{x}_I,y \mid M_I)$ is the likelihood of $M_I$ estimated using the Bayesian Information Criterion (BIC):
\begin{equation}\label{eq2}
\begin{aligned}
&BIC_I = p_I\ln(N)-2\ln{(\Tilde{L}_I)}, \\
&p({\mathbf{x}_I},y \mid M_I) = e^{-BIC_I/2}.
\end{aligned}
\end{equation}

\noindent Here, \(p_I\) is the number of parameters in model \(M_I\), \(N\) is the number of data points in the input data, and \(\Tilde{L}_I\) is the maximized value of the likelihood function for model \(M_I\). Using the above formulas yields:
\begin{equation}\label{eq3}
p(M_I \mid \mathbf{x}_I, y) = \frac{e^{-BIC_I/2}p(M_I)}{\mathcal{L}_{total}},
\end{equation}
where:
\begin{equation}\label{eq3a}
\mathcal{L}_{total} = \sum_{} e^{-BIC/2}p(M),
\end{equation}
%and A satisfies the threshold, \mathcal{L}_{max}, on model likelihood.A %\subseteq\mathcal{P}

In BMA, the Bayes factor plays a crucial role in model selection. Thus, recognizing the relationship between posterior model probabilities and Bayes Factors (BF) application is crucial. While comparing two models, labeled \(I\) and \(J\), we can calculate the Bayes factor for model \(M_I\) against model \(M_J\) using the following expression:
\begin{equation}\label{eq4}
BF_{IJ} = \frac{p(M_I \mid \mathbf{x}_I,y)}{p(M_J \mid \mathbf{x}_J,y)}.
\end{equation}
\noindent If the value of BF exceeds 1, the observed data strongly favors model $M_I$ over model $M_J$. In practical terms, this implies that the information provided by BIC can guide us in selecting the best model that is the model with the highest $log$ of marginal likelihood and, consequently, the smallest BIC. 

Computing output values for the Bayesian averaging logistic regression model involves determining the probability of any predictor variable, $p(x_i)$, and the expected value of the coefficient associated with this predictor variable, i.e. $E[\beta_i]$:
\begin{equation}\label{eq5}
\begin{aligned}
p(x_i) &= \sum_{M_I \ such\ that \ x_i \in \mathbf{x}_I} p(M_I \mid \mathbf{x}_I,y),\\
E[\beta_i] &= \sum_{M_I \ such\ that \ x_i \in \mathbf{x}_I} p(M_I \mid \mathbf{x}_I, y) \times \mathbf{\beta}^I_i.\\
\end{aligned}
\end{equation}
Here, \(i \in \{1, 2, ..., k+1\}\), and \(\beta^I_i\) is the coefficient of \(x_i\) in model \(M_I\). The BayTTA model carries out a series of steps to compute the probabilities: $p(\mathbf{x}) = (p(x_1), p(x_2), ..., p(x_{k+1}))$, and the expected values: \(E[\boldsymbol{\beta}] = (E[\beta_1], E[\beta_2], ..., E[\beta_{k+1}])\). Algorithm 1 presents the pseudocode outlining the main steps of our method.

We define the set \(I_{current} \subseteq \mathcal{P}(\mathbf{x})\) to identify candidate models that we want to process at each iteration. Initially, we form \(I_{current} = \{\{1\}, \{2\}, ..., \{k+1\}\}\), where each \(I \in I_{current}\) corresponds to a model \(M_I\) with one (i.e. \(m=1\)) predictor variable. Subsequently, logistic regression on each model is carried out to calculate the BIC values and the coefficients of the predictor variables. Following this step, we assess the model's likelihood using the estimated value of BIC, as outlined in Eq. \ref{eq2}.  To finalize the model selection process in BMA (see Eq. \ref{eq4}), we specify a uniform prior distribution for all candidate models and set an initial threshold \(\mathcal{L}_{max}\) to zero. 
%where we have the prior distribution being uniform distribution for all candidate models
%If the candidate model likelihood times prior is higher than \(\mathcal{L}_{max}\), we update the threshold \(\mathcal{L}_{max}\) with the model likelihood times prior.} %In this condition, the models considered beforehand and set the maximum likelihood times prior with, \(\mathcal{L}_{max}\).} 
For each model \(M_I\) whose likelihood exceeds \(\mathcal{L}_{max}\), we perform the following steps: (1) replace the threshold \(\mathcal{L}_{max}\) with the likelihood of the model \(M_I\), (2) aggregate its likelihood to calculate \(\mathcal{L}_{total}\) (i.e. the denominator in Eq. \ref{eq3}), and (3) update the probabilities and the coefficients of the predictor variables of the model \(M_I\) using Eqs. \ref{eq5}. At the subsequent iteration, we build a set of candidate models with one additional predictor variable, \(m=m+1\), based on models meeting the threshold at the previous iteration. After repeating this procedure \(k+1\) times, and considering all possible combinations of predictor variables for generating models, the BMA prediction can be calculated as follows:
\begin{equation}\label{eq6}
y_{BMA} = \frac{1}{1+exp(-E[{\boldsymbol{\beta}}]^T\mathbf{x})}.
\end{equation}
In this procedure, the probability of each predictor variable, \(p(x_i)\), corresponds to TTA augmentations, and is defined as the sum of the probabilities of all models that incorporate this predictor variable. Furthermore, the expected value for the coefficient of each predictor variable, \(E[\beta_i]\), is calculated as a weighted average of the coefficients determined by the posterior probability across all models that include this predictor variable. By implementing such a TTA optimization technique that selects more confident augmentations during the testing phase, we ensure that it improves predictive performance, uncertainty estimations, and the overall robustness of deep learning models, compared to simple averaging. 

The uncertainty for the proposed method, $\sigma_{BayTTA}$ is defined based on the mean accuracy values obtained through BMA:

\begin{equation}
\sigma_{BayTTA} = \sqrt{\frac{1}{k+1} \sum_{i=1}^{k+1} \Bigl(p(x_i) \times (acc_i-\mu_{BMA})\Bigr)^2},
\end{equation}
where $acc_i$ is the accuracy of the original image and its $k$ augmented versions obtained using TTA, and $\mu_{BMA}$ is the accuracy obtained using BayTTA.
%\subsection{Bayesian model averaging}

\begin{algorithm}[H]    
\caption{Optimizing TTA using BMA (BayTTA)}\label{Alg}
\begin{algorithmic}[1]
        \Require{Trained model}
        \Require{Original test image $X_{test}$}
        \Require{Set of $k$ transformations}
        \Ensure{Uncertainty estimation for test image $X_{test}$}
        \State $\mathbf{\hat{p}} \leftarrow \mathbf{0}_{k+1}, \hat{E}[\boldsymbol{\beta}] \leftarrow \mathbf{0}_{k+1}$
        \State \(\mathcal{L}_{total} \leftarrow 0, \mathcal{L}_{max} \leftarrow 0\)
        
        \For {$i \leftarrow 1,..., k+1$}
            \State Calculate $I_{next}$ \Comment{$\{I_{next} \subseteq \mathcal{P(\mathbf{x})} \mid \text{length}(S \in I_{next}) = i\}$}
            \State \(I_{current} = \emptyset\)
            \If{\(i == 1\)}
                \State \(I_{current} = I_{next}\)
                \State \(I_{previous} = \emptyset\)
            \Else
                \State  \(A = \{S \in I_{next} \mid S \ \text{contains} \ \text{an element of} \ I_{previous}\}\)
                \State \(I_{current} = I_{current} \cup A \)
                \State \(I_{previous} = \emptyset\)
            \EndIf
            \For{\(I \ \textbf{in} \ I_{current}\)}
                \State $BIC_I, \boldsymbol{\beta}' \leftarrow \text{logits}(\mathbf{x}_I, y)$
                \State $\mathcal{L}_{M_I} \leftarrow e^{-BIC_I/2}$ \Comment{using Eq. \ref{eq2}}
                \If{$\mathcal{L}_{M_I}> \mathcal{L}_{max}$}
                    \State $\mathcal{L}_{total} \mathrel{+}= \mathcal{L}_{M_I}$ 
                    \Comment{using Eq. \ref{eq3a}}
                    \State $\mathcal{L}_{max}= \mathcal{L}_{M_I}$
                    \For{j \ \textbf{in} \ I}
                        \State $\mathbf{\hat{p}}_j \mathrel{+}= \mathcal{L}_{M_I}$ \Comment{for Eqs. \ref{eq5}}
                        \State $\hat{E}[\boldsymbol{\beta}_j] \mathrel{+}= \boldsymbol{\beta'}_j \times \mathcal{L}_{M_I}$ \Comment{for Eqs. \ref{eq5}}
                    \EndFor
                    \State $I_{previous} = I_{previous} \cup I$
                \EndIf
            \EndFor
        \EndFor
    \State Calculate $p(\mathbf{x})$ and $E[\boldsymbol{\beta}]$  \Comment{using Eqs. \ref{eq5}}
    \State Calculate $y_{BMA}$ \Comment{using Eq. \ref{eq6}}
    \State Calculate the uncertainty 
\end{algorithmic}
\end{algorithm}
%\subsection{Bayesian model averaging}
%The uncertainty of our binary classification is

\section{Experimental evaluations and discussion}
In this section, we present the results of our comprehensive experimental study conducted using different DL models in combination with TTA and BayTTA methods. Additionally, we elucidate the experimental setup and the optimization parameters employed in our study. Lastly, we present and discuss the main findings derived from our experimental investigation.

\subsection{Data used in evaluation}
\begin{figure}[!t]
\centering
\includegraphics[scale=.40]{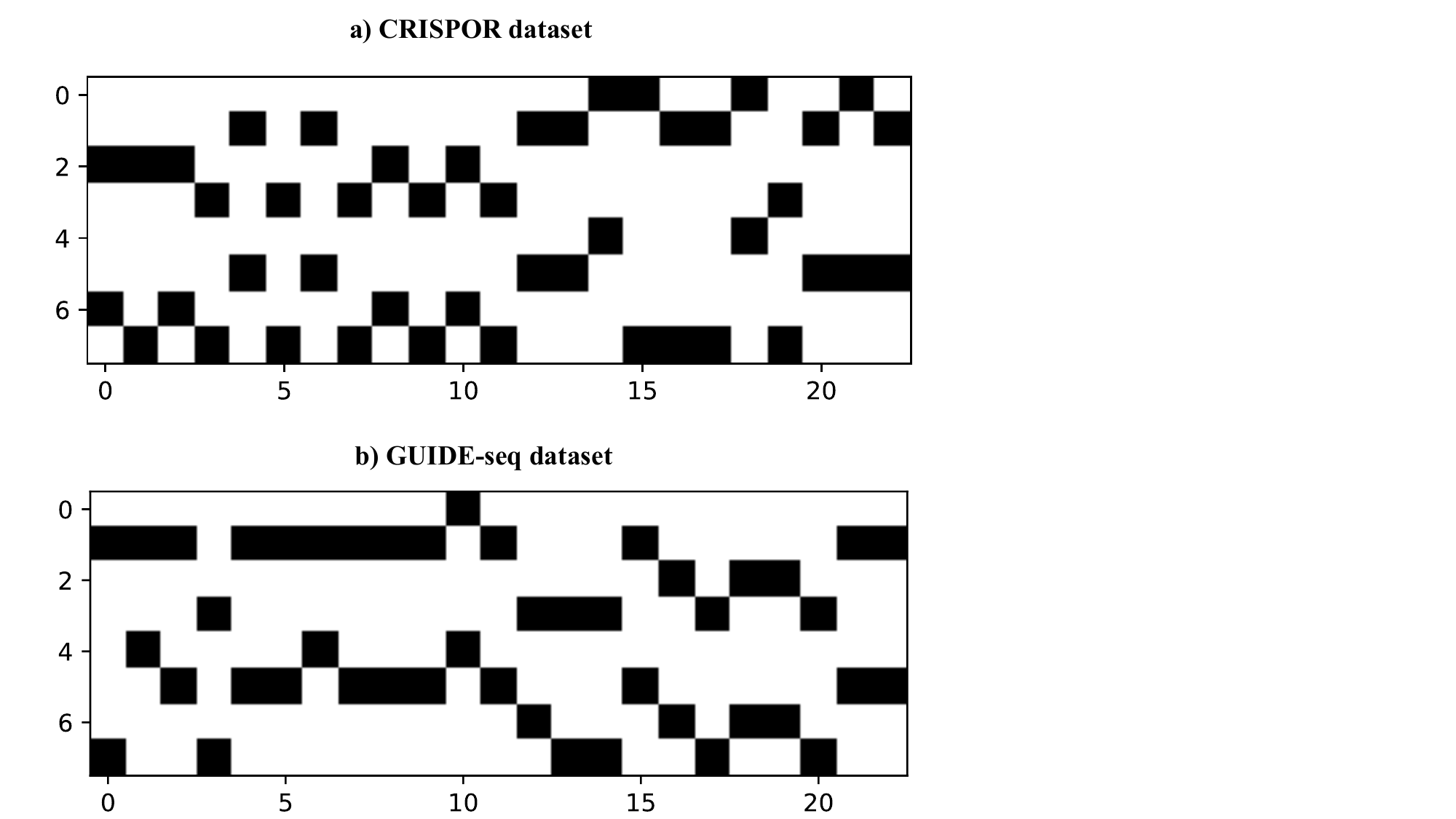}
\caption{Examples of visualizing CRISPR-Cas9 sgRNA-DNA sequence pairs encoded onto 8×23 matrices, then transformed into black and white images from the (a) CRISPOR and (b) GUIDE-seq gene editing datasets, respectively \cite{charlier2021accurate}. These images can be processed by neural networks to predict off-targets generated by CRISPR-Cas9 technology. }\label{fig5}
\end{figure}

We conducted our experiments on three publicly available medical image datasets: skin cancer, breast cancer, and chest X-ray data, which are freely available on the Kaggle and Mendeley data repositories (see also Fig. \ref{fig1}). Subsequently, we evaluated the efficacy of the proposed BayTTA method using two well-known gene editing datasets, CRISPOR and GUIDE-seq, both generated through the CRISPR-Cas-9 technology \cite{sherkatghanad2023using}. We adopted an effective one-hot encoding strategy, originally proposed by Charlier et al. \cite{charlier2021accurate}, to represent sgRNA and DNA sequence pairs as binary images from which off-target effects can be predicted by advanced DL models (see Fig. \ref{fig5}).

\subsubsection{Medical image datasets}
{\bf{Skin cancer dataset}} is the first group of data taken from Kaggle\footnote{https://www.kaggle.com/datasets/fanconic/skin-cancer-malignant-vs-benign}, encompassing images categorized into two distinct classes: Benign and Malignant. This dataset comprises 2637 training images, among which 1440 fall under the Benign category and 1197 under the Malignant category. Additionally, it includes 660 test images, consisting of 360 Benign and 300 Malignant samples. 
%Each image in the dataset is of dimensions $224 \times 224$ pixels.

{\bf{Breast cancer dataset}} is the second group of data supplied by Kaggle\footnote{https://www.kaggle.com/datasets/vuppalaadithyasairam/ultrasound-breast-images-for-breast-cancer}, comprising 8116 training images. This dataset has 4074 samples categorized as Benign and 4042 samples categorized as Malignant. Additionally, the dataset includes 900 test samples, 500 of which are classified as Benign and 400 as Malignant. 
%Each image in the dataset is of dimensions $224 \times 224$ pixels.

{\bf{Chest X-ray dataset}} is a large collection of X-ray images extracted from a publicly available medical image database\footnote{https://easy.dans.knaw.nl/ui/datasets/id/easy-dataset:102451} \cite{kermany2018large}. The dataset contains 5216 training images, among which 1341 are categorized as Normal and 3875 as Pneumonia samples. Additionally, the test dataset includes 624 images, among which 234 are classified as Normal and 390 as Pneumonia samples. 
%Each image in the dataset is of dimensions $224 \times 224$ pixels.

\subsubsection{Gene editing datasets}

{\bf{CRISPOR}} database, organized and maintained by Maximilian Haeussler \cite{haeussler2016evaluation} aggregates widely-used public datasets aimed at quantifying on-target guide efficiency and detecting off-target cleavage sites\footnote{http://crispor.tefor.net}. The dataset we selected from this database (see also \cite{charlier2021accurate}) comprises $18,211$ black and white training images (each with $8 \times 23$ pixel dimension), among which 18,112 are categorized as on-targets and 99 as off-targets. Additionally, the dataset includes 7806 testing images of the same dimension, 7763 of which are classified as on-targets and 43 as off-targets.

{\bf{GUIDE-seq}} is one of the pioneering off-target data repositories, derived from the outcomes of the GUIDE-seq technique developed by Tsai et al. \cite{tsai2015guide}. It serves as an accurate framework for genome-wide identification of off-target effects. The sgRNAs used in GUIDE-seq target the following sites: VEGFA site 1, VEGFA site 2, VEGFA site 3, FANCF, HEK293 site 2, HEK293 site 3, and HEK293 site 4, wherein 28 off-targets with a minimum modification frequency of $0.1$ were identified (among 403 potential off-targets). This dataset consists of black and white images with $8 \times 23$ pixel dimension. It comprises 309 training images, including 291 on-target and 18 off-target samples. Additionally, it comprises 133 testing images, including 121 on-target and 12 off-target samples.

\begin{table*}[h!]
\caption{\label{tab1} Hyperparameter configuration of the pre-trained deep learning models (VGG-16, MobileNetV2, DenseNet201, ResNet152V2, and InceptionResNetV2) for the Skin Cancer, Breast Cancer, and Chest X-ray medical image datasets considered in our study.}
\centering
\begin{tabular}{lcccccccccccc}
\toprule
\multicolumn{1}{c}{} & \multicolumn{4}{c}{Skin Cancer} & \multicolumn{4}{c}{Breast Cancer}&\multicolumn{4}{c}{Chest X-ray} \\
\cmidrule(rl){2-5} \cmidrule(rl){6-9} \cmidrule(rl){10-13}
{Models} & {Opt} & {LR} & {BS} &{Epoch} &  {Opt} & {LR} & {BS} & {Epoch} &{Opt} & {LR} & {BS} & {Epoch} \\
\midrule
VGG-16&Adam&0.0005&256&124&Adamax&0.005&128&157&Adamax&0.0001&128&156\\ 
\midrule                   
MobileNetV2&Adam&0.001&256&82&SGD&0.001&16&171&SGD&0.00001&256&100\\ 
\midrule                            
DenseNet201&Adamax&0.0001&32&138&SGD&0.001&16&75&SGD&0.0001&256&145\\ 
\midrule                            
ResNet152V2&Adamax&0.0005&128&127&SGD&0.001&16&251&SGD&0.00005&256&200\\
\midrule 
Inception&Adamax&0.0001&16&256&SGD&0.001&16&249&SGD&0.00001&256&140\\
ResNetV2&&&&&&&&&&\\ 
\bottomrule
\end{tabular}
\end{table*}

\subsection{Implementation details and model settings}
To facilitate a comprehensive understanding of the proposed BayTTA method, we provide additional insights into the selected experimental configuration.

\indent {\bf{Experimental Setup.}} For the training phase, we used a Compute Canada cluster equipped with NVIDIA Tesla P100 and NVIDIA v100 GPUs (referred to as the Cedar cluster). Additionally, we used a software environment with Python, TensorFlow, and PyTorch stack.

{\bf{Pre-trained CNN models.}} During the training phase of our first experiment, we employed five well-established convolutional networks (CNNs), namely VGG-16, MobileNetV2, DenseNet201, ResNet152V2, and InceptionResNetV2, as our feature extraction backbone, initializing weights through pre-training on the ImageNet dataset. The baseline versions of these models were initially used (without applying TTA or BayTTA), followed by their application in combination with TTA and BayTTA. All medical images were resized to the $224 \times 224$ pixel size to ensure uniformity in model inputs. The hyperparameters, including the optimizer (Opt), the learning rate (LR), the batch size (BS), and the number of epochs for each of the five pre-trained models and each medical image dataset are summarized in Table \ref{tab1}. Additionally, we employed the early stopping technique to handle overfitting \cite{prechelt2002early}. During our network training, the data augmentation techniques were implemented using the following parameters: rotation range = 20, width shift range = 0.2, height shift range = 0.2, shear range = 0.2, zoom range = 0.2, horizontal flip, and vertical flip.

Four pre-trained CNN models, VGG-16, MobileNetV2, DenseNet201, and ResNet152V2, as well as a custom CNN with five layers, comprising two convolutional and three fully connected dense layers, were used to predict off-targets in highly imbalanced gene editing datasets.
The models were trained using the RMSprop optimizer, with a learning rate of $0.001$ and a batch size of $64$. To address potential overfitting issues a set of callback functions was used during training, specifically the ReduceLROnPlateau and EarlyStopping functions.

%. This optimizer was chosen for its effectiveness with such imbalanced datasets, and 
%To assess the model's performance, we computed several commonly used metrics, including precision, recall, F1-score, and accuracy, and we also examined the Receiver Operator Characteristic (ROC) along with its Area Under the Curve (AUC).

%\begin{table*}
%\caption{\label{tab7} Accuracy results of our proposed method on the CRISPOR and GUIDE-seq datasets.}
%\centering
%\begin{tabular}{lcc}
%\toprule
%Models & CRISPOR & GUIDE-seq\\
%\midrule
%CNN&99.82 $\pm$ 0.02 &90.72 $\pm$ 0.36\\ 
%CNN+TTA&99.86 $\pm$ 0.008&91.35 $\pm$ 0.35 \\ 
%\bf{CNN+BayTTA}&\bf{99.87$\pm$ 0.008} &\bf{91.73 $\pm$ 0.16}\\   
%\bottomrule
%\end{tabular}
%\end{table*}

{\bf{State-of-the-art DL models intended for medical image analysis.}} 
%\subsection{State-of-the-art DL models}
 To evaluate the performance of BayTTA, we also compared its performance against TTA and baseline, considering some state-of-the-art DL models recently used in medical image analysis, namely Attention Residual Learning (ARL) \cite{zhang2019attention}, COVID-19 \cite{ozturk2020automated}, LoTeNet \cite{raghav2020tensor}, IRv2+SA \cite{datta2021soft}, and PCXRNet \cite{feng2022pcxrnet}. 
%It's important to highlight that all the state-of-the-art methods we compare in this context also utilize the same.

IRv2+SA \cite{datta2021soft} explores the efficacy of soft attention in deep neural architectures. The core objective of this technique is to emphasize the significance of essential features, while mitigating the influence of noise-inducing ones. The ARL model, proposed by Zhang et al. \cite{zhang2019attention}, is intended for classifying skin lesions in dermoscopy images. The model architecture includes several ARL blocks, a global average pooling layer, and a classification layer. Each ARL block employs a combination of residual learning and an innovative attention-learning mechanism to boost the capacity of discriminative representations. Ozturk et al. \cite{ozturk2020automated} proposed a novel model based on the DarkNet method designed for automated detection of COVID-19 cases using chest X-ray images to deliver precise diagnostic results in the framework of binary and multi-class classifications. PCXRNet \cite{feng2022pcxrnet} is an attention-driven convolutional neural network designed for pneumonia diagnosis based on chest X-ray image analysis. PCXRNet incorporates a novel Condensed Attention Module (CDSE) to harness the information within the feature map channels. The LoTeNet (Locally order-less Tensor Network) model \cite{raghav2020tensor} is based on the use of tensor networks, a crucial tool for physicists to analyze complex quantum many-body systems.

\subsection{Experimental results}
%{\textcolor{red} {To explore the generalization ability of the BayTTA model, we apply it to different medical image datasets and gene editing datasets for classification task. First, we use five individual models as the baseline methods for implementation on medical image datasets. Then, we apply the BayTTA model on the baseline methods with highest accuracy.}}

\begin{table*}[!t]
\caption{\label{tab2} Comparison of the baseline CNN model accuracy (\%) $\pm$ STD performance against the TTA and BayTTA versions on the skin cancer dataset. The highest accuracy per column is in bold. The asterisk (\(*\)) denotes the highest overall accuracy obtained by the models.}
\centering
\begin{tabular}{lccccc}
\toprule
Models & VGG-16 & MobileNetV2 & DenseNet201& ResNet152V2&InceptionResNetV2\\
\hline
Baseline&84.95 $\pm$ 0.40&85.75 $\pm$ 1.31&88.28 $\pm$ 0.76&83.33 $\pm$ 1.75&81.63 $\pm$ 1.70\\ 
TTA&85.24 $\pm$ 0.39 &87.39 $\pm$ 0.42 &88.33 $\pm$ 0.58 &83.82 $\pm$ 0.52&83.01 $\pm$ 0.77 \\ 
\bf{BayTTA}&\bf{85.50 $\pm$ 0.14} &\bf{87.52 $\pm$ 0.11} &\bf{89.38 $\pm$ 0.17*} &\bf{84.04 $\pm$ 0.09} &\bf{83.98 $\pm$ 0.46} \\ 
\bottomrule
\end{tabular}

\bigskip 

\caption{\label{tab3} Comparison of the baseline CNN model accuracy (\%) $\pm$ STD performance against the TTA and BayTTA versions on the breast cancer dataset. The highest accuracy per column is in bold. The asterisk (\(*\)) denotes the highest overall accuracy obtained by the models.}
\centering
\begin{tabular}{lccccc}
\toprule
Models & VGG-16 & MobileNetV2 & DenseNet201& ResNet152V2&InceptionResNetV2\\
\hline
Baseline&88.92 $\pm$ 1.70&86.70 $\pm$ 0.94&\bf{86.81 $\pm$ 1.06}&91.52 $\pm$ 1.18&91.25 $\pm$ 0.98\\ 
TTA&89.64 $\pm$ 0.52 &87.84 $\pm$ 0.81 &86.64 $\pm$ 0.74 &93.64 $\pm$ 1.04 &\bf{93.13 $\pm$ 0.71}\\ 
\bf{BayTTA}&\bf{90.11 $\pm$ 0.64}&\bf{88.36 $\pm$ 0.34} &86.18 $\pm$ 0.59&\bf{93.81 $\pm$ 0.99*} &92.84 $\pm$ 0.69\\ 
\bottomrule
\end{tabular}
\bigskip
\caption{\label{tab4} Comparison of the baseline CNN model accuracy (\%) $\pm$ STD performance against the TTA and BayTTA versions on the chest X-ray dataset. The highest accuracy per column is in bold. The asterisk (\(*\)) denotes the highest overall accuracy obtained by the models.}
\centering
\begin{tabular}{lccccc}
\toprule
Models & VGG-16 & MobileNetV2 & DenseNet201& ResNet152V2&InceptionResNetV2\\
\hline
Baseline&71.02 $\pm$ 1.01&61.53 $\pm$ 0.73&66.77 $\pm$ 1.11&62.45 $\pm$ 0.17&63.31 $\pm$ 0.57\\ 
TTA&71.11 $\pm$ 0.78 &61.32 $\pm$ 0.62 &68.20 $\pm$ 0.63 &62.54 $\pm$ 0.19 &63.45 $\pm$ 0.56\\ 
\bf{BayTTA}&\bf{72.49 $\pm$ 0.25*} &\bf{62.50 $\pm$ 0.27} &\bf{69.98 $\pm$ 0.28}&\bf{62.82 $\pm$ 0.06} &\bf{64.30 $\pm$ 0.21}\\ 
\bottomrule
\end{tabular}
\end{table*}

\begin{table*}[h!]
\caption{\label{tab5} Comparison of state-of-the-art classification models against their TTA and BayTTA counterparts, in terms of accuracy (\%) and STD, after their application on the skin cancer, breast cancer, and chest X-ray datasets considered in our study. The highest overall accuracy per dataset is highlighted in bold.} %The asterisk denotes the superior performance of BayTTA compared to TTA.}
\centering
\begin{tabular}{llll}
\toprule
Models & Skin Cancer & Breast Cancer & Chest X-ray\\
\hline
ARL \cite{zhang2019attention}&86.21 $\pm$ 0.42  &84.87 $\pm$ 0.15&64.58 $\pm$ 0.15\\

COVID-19 \cite{ozturk2020automated}& 85.65 $\pm$ 0.61 &	94.92 $\pm$ 0.88 &	84.07 $\pm$ 1.41\\

LoTeNet \cite{raghav2020tensor}&74.89 $\pm$ 0.72  &72.63  $\pm$ 1.24 &79.48 $\pm$  0.80\\     

IRv2+SA \cite{datta2021soft}&90.92 $\pm$ 0.32  &95.84 $\pm$ 0.54 &\bf{87.76} $\pm$ 1.20 \\
                                               
PCXRNet \cite{feng2022pcxrnet}&79.70 $\pm$ 0.21  &93.05 $\pm$ 0.48 &79.15 $\pm$ 1.46\\
 
\bf{COVID-19+TTA}&85.70 $\pm$ 0.74	&94.88 $\pm$ 0.34	&79.10 $\pm$ 1.96\\ 
                            
\bf{COVID-19+BayTTA}&87.17 $\pm$ 0.31	&95.55 $\pm$ 0.31	&85.04 $\pm$ 1.43\\ 
                                                     
\bf{IRv2+SA+TTA}&89.74 $\pm$ 0.71 &94.48 $\pm$ 0.67&83.71 $\pm$ 1.36\\ 
                            
\bf{IRv2+SA+BayTTA}&\bf{91.07 $\pm$ 0.04}  &\bf{96.66 $\pm$ 0.49}&\bf{87.76 $\pm$ 1.20}\\
\bottomrule
\end{tabular}
\bigskip 
\caption{\label{tab6} Comparison of state-of-the-art classification models against their TTA and BayTTA counterparts, in terms of precision (PR (\%)), recall (RE (\%)), and F1-score (FS (\%)), after their application on the skin cancer, breast cancer, and chest X-ray datasets considered in our study. The highest overall accuracy per dataset is highlighted in bold.}
\centering
\begin{tabular}{lccccccccc}
\toprule
\multicolumn{1}{l}{} & \multicolumn{3}{c}{Skin Cancer} & \multicolumn{3}{c}{Breast Cancer}&\multicolumn{3}{c}{Chest X-ray} \\
\cmidrule(rl){2-4} \cmidrule(rl){5-7} \cmidrule(rl){8-10}
{Models} &  {PR} & {RE} &{FS} &{PR} & {RE} &{FS} & {PR} & {RE} &{FS} \\
\hline
                   
ARL \cite{zhang2019attention}&86.66&85.80&86.09&89.60&83.15&84.33&81.46&51.46&41.57\\
                           
COVID-19 \cite{ozturk2020automated}&84.78&83.44&84.32&94.29&94.42&94.29&80.24&98.97&88.61\\
                           
LoTeNet \cite{raghav2020tensor}&75.38&86.50&66.80&75.22&62.25&68.13&77.53&96.41&85.94\\ 
   
IRv2+SA \cite{datta2021soft}&\bf{89.60}&90.55&90.04&\bf{97.68}&95.16&95.52&84.24&98.97&\bf{91.03}\\
 
PCXRNet \cite{feng2022pcxrnet}&79.98&\bf{99.51}&88.69&93.03&73.90&82.37&75.22&\bf{99.37}&85.63\\
 
\bf{COVID-19+TTA}&83.81&84.61&84.17&94.55&94.05&94.26&\bf{99.17}&75.67&85.61\\ 
                            
\bf{COVID-19+BayTTA}&85.80&86.01&85.90&94.41&95.75&95.05&81.31&98.80&89.53\\ 
                           
\bf{IRv2+SA+TTA}&87.22&84.63&90.37&98.59&91.58&95.19&79.92&99.57&88.66\\ 
                           
\bf{IRv2+SA+BayTTA}&88.70&92.11&\bf{90.42}&95.99&\bf{96.58}&\bf{96.25}&84.24&98.97&\bf{91.03}\\ 
\bottomrule
\end{tabular}
\end{table*}

\subsubsection{Results for medical image datasets}
We present a thorough experimental assessment of the BayTTA model in two stages. 

First, we evaluate its performance in the case when it was used in combination with five pre-trained CNN models, including VGG-16, MobileNetV2, DenseNet201, ResNet152V2, and InceptionResNetV2, to determine the most suitable backbone model for the skin cancer (Table \ref{tab2}), breast cancer (Table \ref{tab3}), and chest X-ray (Table \ref{tab4}) datasets. In this study, we executed the baseline models three times, computing the mean accuracy and the standard deviation (STD) of accuracy. We applied TTA during the testing phase on the baseline models to assess the model accuracy and conducted a comparative analysis with BayTTA, in which the BMA method was used for model averaging. We considered each original image and six of its random augmentations (obtained through rotations) during the testing phase, evaluating the mean accuracy and STD obtained with these random augmentations. 
The results reported in Tables \ref{tab2} demonstrate that the use of BayTTA allowed us to outperform the baseline and TTA-based models, achieving the highest accuracy and significantly reducing the standard deviation values for the skin cancer dataset. Regarding the breast cancer dataset, BayTTA outperforms the baseline and TTA-based models in accuracy for the three out of five pre-trained CNNs, i.e. VGG-16, MobileNetV2, and ResNet152V2. Nonetheless, the BayTTA-based networks consistently exhibited the lowest STD values, as outlined in Table \ref{tab3}. In the case of the imbalanced chest X-ray dataset (see Table \ref{tab4}), the combination of TTA and BMA (i.e. BayTTA) allowed us to improve the baseline and TTA results in all cases in terms of both accuracy and STD.

\begin{table*}[h!]
\caption{\label{tab7} Comparison of the baseline CNN model accuracy (\%) $\pm$ STD performance against their TTA and BayTTA versions on the CRISPOR dataset. The highest accuracy per column is in bold. The asterisk (\(*\)) denotes the highest overall accuracy obtained by the models.}
\centering
\begin{tabular}{lccccc}
\toprule
Models & VGG-16 & MobileNetV2 & DenseNet201& ResNet152V2& CNN-5 layers\\
\hline
Baseline&\bf{99.41 $\pm$ 0.005}&\bf{99.55} $\pm$0.04&\bf{99.53} $\pm$ 0.05&\bf{99.62} $\pm$ 0.08&99.82 $\pm$ 0.057\\ 
TTA&99.37 $\pm$ 0.018&99.46 $\pm$ 0.05&99.46 $\pm$ 0.03&99.53 $\pm$ 0.04&99.77 $\pm$ 0.016\\ 
\bf{BayTTA}&\bf{99.41} $\pm$ 0.011&\bf{99.55 $\pm$ 0.04}&\bf{99.53 $\pm$ 0.03}&\bf{99.62 $\pm$ 0.02}&\bf{99.84 $\pm$ 0.008*}\\ 
\bottomrule
\end{tabular}
\bigskip
\caption{\label{tab8} Comparison of the baseline CNN model accuracy (\%) $\pm$ STD performance against their TTA and BayTTA versions on the GUIDE-seq dataset. The highest accuracy per column is in bold. The asterisk (\(*\)) denotes the highest overall accuracy obtained by the models.}
\centering
\begin{tabular}{lccccc}
\toprule
Models & VGG-16 & MobileNetV2 & DenseNet201& ResNet152V2& CNN-5 layers\\
\hline
Baseline&\bf{93.51} $\pm$ 0.98& 90.47 $\pm$ 0.70&90.97 $\pm$ 0.60&87.95 $\pm$ 0.02&94.22 $\pm$ 0.92\\ 
TTA&91.61 $\pm$ 1.18&90.54 $\pm$ 0.55&90.83 $\pm$ 0.50&90.65 $\pm$ 0.79&94.45 $\pm$ 0.26\\ 
\bf{BayTTA}&\bf{93.51 $\pm$ 0.67}&\bf{90.97$\pm$ 0.15}&\bf{91.72 $\pm$ 0.30}&\bf{90.98 $\pm$ 0.16}&\bf{94.73 $\pm$ 0.11*}\\ 
\bottomrule
\end{tabular}
\bigskip 
\caption{\label{tab9} Comparison of state-of-the-art classification models against their TTA and BayTTA counterparts, in terms of accuracy (\%) and STD on the CRISPOR and GUIDE-seq gene editing datasets. The highest accuracy per column is highlighted in bold.}%The asterisk denotes the superior performance of BayTTA compared to TTA.}
\centering
\begin{tabular}{lll}
\toprule
Models &  CRISPOR & GUIDE-seq\\
\hline
ARL \cite{zhang2019attention}&94.20 $\pm$ 1.19 & 91.04 $\pm$ 0.31 \\

COVID-19 \cite{ozturk2020automated}& 99.50 $\pm$ 0.20 & 93.87 $\pm$ 0.96 \\

LoTeNet \cite{raghav2020tensor}&98.85 $\pm$ 0.43&93.38 $\pm$ 0.99\\     

IRv2+SA \cite{datta2021soft}&99.38 $\pm$ 0.24 & 91.22 $\pm$ 1.03 \\
                                              
PCXRNet \cite{feng2022pcxrnet}&95.69 $\pm$ 2.51 & 91.42 $\pm$ 0.97 \\
 
\bf{COVID-19+TTA}&99.39 $\pm$ 0.18 & 95.07 $\pm$ 0.90\\ 
                            
\bf{COVID-19+BayTTA}&\bf{99.66} $\pm$ 0.02 & \bf{96.56} $\pm$ 0.41\\ 
                                                     
\bf{IRv2+SA+TTA}&95.24 $\pm$ 0.57& 93.75 $\pm$ 0.71\\ 
                            
\bf{IRv2+SA+BayTTA}&99.43 $\pm$ 0.38& 94.29 $\pm$ 0.43\\
\bottomrule
\end{tabular}
\bigskip
\caption{\label{tab10} 
Comparison of state-of-the-art classification models against their TTA and BayTTA counterparts, in terms of precision (PR (\%)), recall (RE (\%)), and F1-score (FS (\%)) on the CRISPOR and GUIDE-seq gene editing datasets. The highest accuracy per column is highlighted in bold.}
\centering
\begin{tabular}{lcccccc}
\toprule
\multicolumn{1}{l}{} & \multicolumn{3}{c}{CRISPOR} & \multicolumn{3}{c}{GUIDE-seq} \\
\cmidrule(rl){2-4} \cmidrule(rl){5-7}
{Models} &  {PR} & {RE} &{FS}  & {PR} & {RE} & {FS} \\
\hline
                  
ARL \cite{zhang2019attention}&88.33&	93.66& 90.72&33.91&56.81&42.38\\
                           
COVID-19 \cite{ozturk2020automated}&98.03&98.07&98.04&20.73&	33.30& 37.93\\
                           
LoTeNet \cite{raghav2020tensor}&89.74&92.10&94.59&33.30&12.12&17.74\\ 
   
IRv2+SA \cite{datta2021soft}&	\bf{99.73}&	95.20&96.71&	37.86&	41.91&37.54\\
 
PCXRNet \cite{feng2022pcxrnet}&	74.13&	98.01&84.43&	32.90&	69.67& 43.47\\
 
\bf{COVID-19+TTA}&	96.12&	97.41&97.55&	60.68&	60.69&58.78 \\ 
                            
\bf{COVID-19+BayTTA}&	99.36&	\bf{98.08}&\bf{98.70}&\bf{80.79}&	\bf{70.03}& \bf{72.64}\\ 
                            
\bf{IRv2+SA+TTA}&	88.99&	95.91&91.92 &	30.76&	7.76&10.18 \\ 
                           
\bf{IRv2+SA+BayTTA}&	\bf{99.73}&95.61&97.49&	50.01&	8.61&14.54 \\ 
\bottomrule
\end{tabular}
\end{table*}

Second, we conducted a comparative study of the TTA-based, BayTTA-based, and baseline state-of-the-art classification models which have been recently used in the literature to process medical image data (see Tables \ref{tab5} and \ref{tab6}). To ensure a thorough performance assessment, we reported the experimental results for the four following key metrics: accuracy, precision, recall, and F1-score. Our analysis revealed that among state-of-the-art models compared, COVID-19 and IRv2+SA provided the highest overall accuracy, surpassing the results of ARL, LoTeNet, and PCXRNet models. Thus, we used the TTA and BayTTA frameworks in combination with these two top-performing models. In this experiment, we considered original images as well as ten random augmentation samples generated for each of them using rotations during the test phase. Our results, reported in Table \ref{tab5}, indicate that the use of TTA leads to a decrease in the accuracy of IRv2+SA, but the performance of TTA is notably worse for the imbalanced chest X-ray dataset. Nonetheless, BayTTA improves the models accuracy even in cases of suboptimal performance of TTA, demonstrating its high effectiveness. The model performances in terms of precision, recall,
and F1-score (see Table \ref{tab6}) also demonstrate a much higher robustness of the proposed BayTTA computational framework, compared to the TTA-based and baseline models.

\subsubsection{Results for gene editing datasets}
Similarly to medical image data, we first evaluated the performance of the proposed BayTTA method on gene editing datasets using it in combination with four pre-trained CNN models VGG-16, MobileNetV2, DenseNet201, ResNet152V2 as well as a custom CNN model with 5 layers. A comparison against the TTA-based and baseline models was carried out. The results obtained are reported in Tables \ref{tab7} and \ref{tab8}. 

We executed each baseline model three times, computing the mean accuracy and standard deviation (STD) of accuracy. During the testing phase, we used original images along with six random augmented samples (generated for each original image through rotations and random erasing). By observing the results obtained for both CRISPOR and GUIDE-seq gene editing datasets, we can conclude that BayTTA demonstrated an enhanced accuracy performance, while consistently yielding lower standard deviation values.

\begin{figure*}
\centering
\includegraphics[scale=.5]{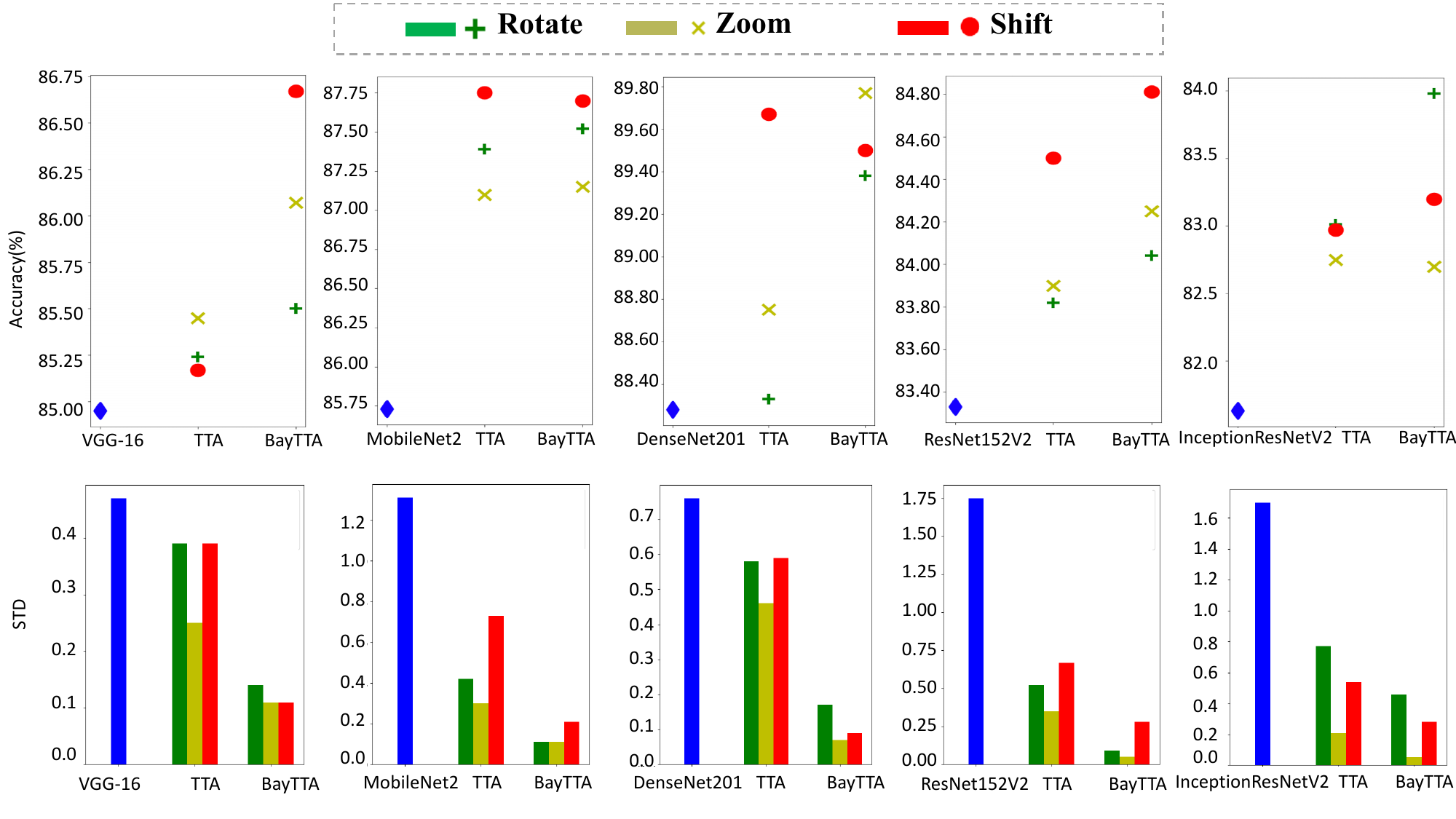}
\caption{Comparison of the TTA and BayTTA method performance on the skin cancer dataset in terms of accuracy and standard deviation, while considering pre-trained baseline models with rotate, zoom, and shift augmentations. }\label{fig7}
\end{figure*}
\begin{figure*}
\centering
\includegraphics[scale=.5]{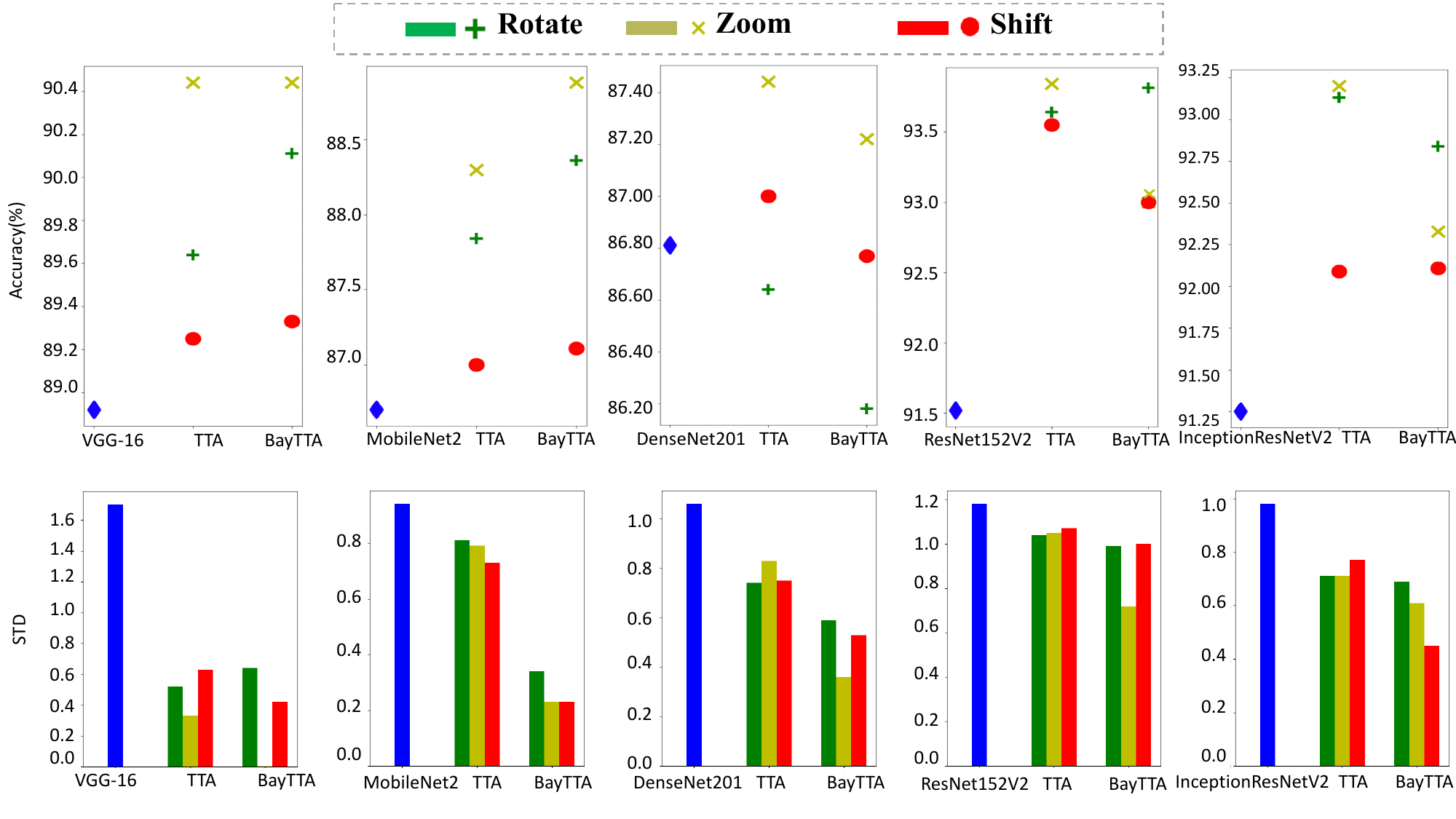}
\caption{Comparison of the TTA and BayTTA method performance on the breast cancer dataset in terms of accuracy and standard deviation, while considering the pre-trained baseline models with rotate, zoom, and shift augmentations.}\label{fig8}
\end{figure*}
\begin{figure*}
\centering
\includegraphics[scale=.5]{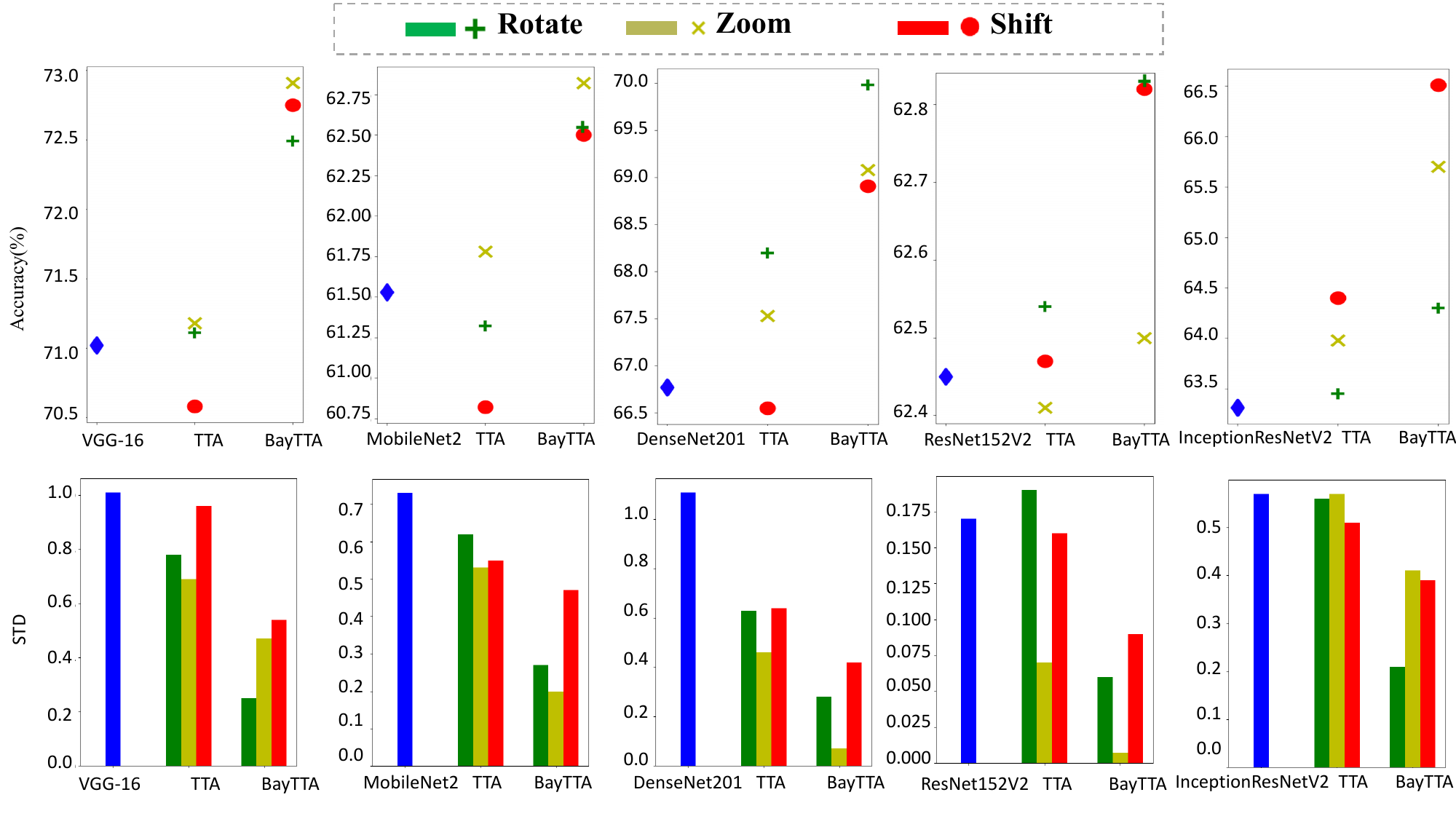}
\caption{Comparison of the TTA and BayTTA method performance on the chest X-ray dataset in terms of accuracy and standard deviation, while considering the pre-trained baseline models with rotate, zoom, and shift augmentations.}\label{fig9}
\end{figure*}

In addition to the comparison with pre-trained baseline models, we conducted a comparative analysis between BayTTA and TTA used in combination with state-of-the-art DL classification models whose input was adapted to gene editing datasets. The results obtained in this experiment are presented in Tables \ref{tab9} and \ref{tab10}. To conduct our assessment, we used original images with ten randomly created samples (generated for each original image during the testing phase using rotations and random erasing). The results reported in Table \ref{tab9} suggest that the proposed BayTTA computational framework used in combination with the COVID-19 model provided the highest accuracy and the lowest STD values for both CRISPOR and GUIDE-seq datasets. Similar model performances can be observed in Table \ref{tab10}, where the obtained precision, recall, and F1-score metric values are reported - the COVID-19+BayTTA model provided the best results overall.

\subsection{Assessing the impact of different data augmentations and of increasing the number of samples}

Furthermore, we conducted a number of supplementary experiments to determine the key factors that may influence the performance of the TTA and BayTTA methods considered in this study. Our analysis was performed on the above-described skin cancer, breast cancer, and chest X-ray datasets to study the influence of sample sizes and different data augmentation techniques and sample sizes.

\subsection{Evaluation of different data augmentations}

To analyze the impact of various data augmentation types and compare the results with traditional TTA, we tested the performance of the proposed BayTTA method on the three following augmentation types: rotation, zoom, and shift. These augmentations were incorporated into the data augmentation process of TTA and BayTTA at the testing phase. During testing, we used six augmented samples in addition to the original image in our experiments including the VGG-16, MobileNetV2, DenseNet201, ResNet152V2, and InceptionResNetV2 pre-trained CNN models.

Our results are graphically represented in Figs. \ref{fig7}, \ref{fig8}, and \ref{fig9}, corresponding to the skin cancer, breast cancer, and chest X-ray datasets, respectively. These graphs demonstrate how the accuracy and STD of the model alter as we vary data augmentation types, highlighting which augmentations offer higher accuracy and boost the robustness of CNN models.

We can conclude that BayTTA generally enhances classification performance by increasing the models accuracy and reducing STD after integrating various data augmentations. This outcome was expected, as our proposed method combines predictions from each candidate model based on a constraint on model likelihood obtained from logistic regression. This combination results in model-averaged predictions that account for both the model's uncertainty and accuracy.
\begin{figure*}
\centering
\includegraphics[scale=.29]{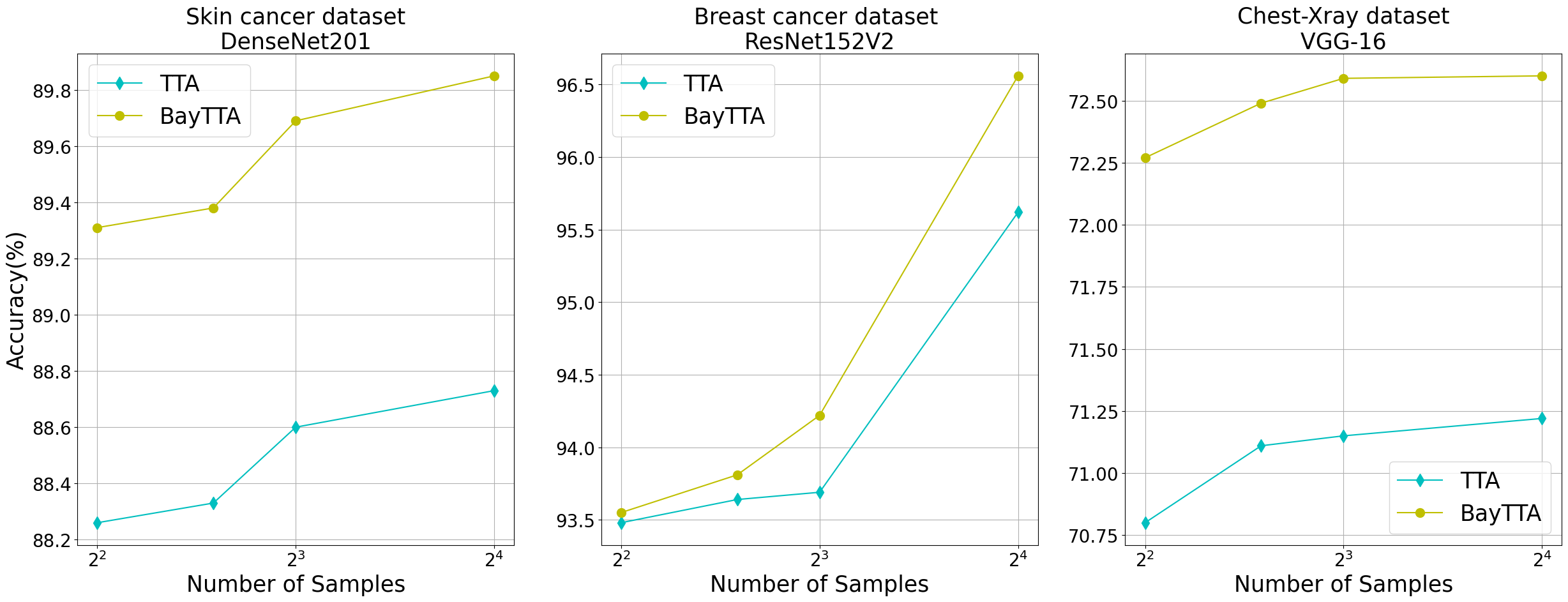}
\caption{Comparison of the TTA and BayTTA method performance in terms of accuracy with different numbers of samples, while considering the best baseline model from Tables \ref{tab2}, \ref{tab3}, and \ref{tab4} for each medical image dataset.}\label{fig10}
\end{figure*}

\subsection{Evaluation of increasing the number of samples}

We analyzed the impact of increasing the number of augmented samples on the models accuracy during the testing phase. As indicated in Tables \ref{tab2}, \ref{tab3} and \ref{tab4}, the combination of BMA and TTA contributes to a higher accuracy of DenseNet201 on the skin cancer dataset, of ResNet152V2 on the breast cancer dataset, and of VGG-16 on the chest X-ray dataset, compared to baseline models and TTA. Thus, we examined the effect of increasing the number of augmented samples in these three specific cases. 

Fig. \ref{fig10} demonstrates that the accuracy of both TTA and BayTTA increases as the number of samples in the test data with diverse transformations grows. Hence, training the model to recognize patterns in their diverse forms generally enhances its ability to make accurate predictions on unseen examples. Obviously, the increase in the amount of augmented data results in a higher accuracy at the expense of a higher computational cost.

\section{Conclusion} 
In this study, we investigated how the combination of Test-Time Augmentation (TTA) and Bayesian Model Averaging (BMA) techniques can improve the accuracy of pre-trained and state-of-the-art deep learning models applied in the field of medical image classification. In particular, we focused on the two following modeling aspects: TTA optimization and uncertainty evaluation based on posterior probabilities. Our empirical observations indicate that using BMA as a method for combining model predictions is highly effective for enhancing the performance of traditional TTA. BMA allows one to assign weights to different models based on their posterior probabilities. Therefore, the proposed BayTTA technique can be viewed as a novel and effective methodology that harnesses the combined strengths of TTA and BMA. One of the key strengths of our approach is its capacity to quantify model uncertainty, which means that we not only obtain more accurate predictions, but also gain insight into the level of confidence the model has in each of its predictions. This insight is crucial, especially in applications where incorrect predictions can lead to critical consequences, such as medical image and gene editing analyses. 

%{\color{red} Our results suggest an interesting outcome:
%\begin{itemize}
%\item Incorporating both the standard TTA and the proposed BayTTA technique into the IRv2+SA \cite{datta2021soft} was motivated by the superior overall performance of this state-of-the-art deep learning model compared to others. In this case, standard TTA leads to a decrease in the accuracy of IRv2+SA, but proposed BayTTA technique compensate the results and improve both accuracy and STD values. The proposed technique even outperforms the IRv2+SA models and lead to higher value of accuracy
%\item For breast cancer dataset, the model is more accurate, and we achieve the lower TTA gain. 
%\item The performance of TTA is notably worse for the imbalanced chest X-ray dataset compared
%to the other two datasets.
%\end{itemize}}
%. We employ BMA in this study to aggregate the accuracy results obtained through TTA on medical image datasets. 

\section{Acknowledgments}
This work was supported by le Fonds Québécois de la Recherche sur la Nature et les Technologies [grant 173878] and the Natural Sciences and Engineering Research Council of Canada [grant 249644].
%%Harvard

%\bibliographystyle{unsrt}
%\bibliography{refs}

\end{document}